\documentclass[runningheads]{llncs}
\usepackage{graphicx}
\usepackage{amsmath,amssymb} % define this before the line numbering.
\usepackage{color}
\usepackage{subcaption}

\begin{document}
\pagestyle{headings}
\mainmatter

\def\ACCV22SubNumber{748}  % Insert your submission number here

%===========================================================
\title{Understanding the Impact of Image Quality and Distance of Objects to Object Detection Performance} % Replace with your title
\titlerunning{Understanding the Impact}
\authorrunning{Y. Hao et al.}

\author{ Yu Hao\inst{1,2}\and
Haoyang Pei\inst{2}\and
Yixuan Lyu\inst{2}\and
Zhongzheng Yuan\inst{2} \and
John-Ross Rizzo\inst{2,4} \and
Yao Wang\inst{2} \and
Yi Fang\inst{1,2,3}}
\institute{NYU Multimedia and Visual Computing Lab\\
NYU Tandon School of Engineering, New York University, USA\\
New York University Abu Dhabi, UAE\\
NYU Langone Health, USA}

\maketitle

%===========================================================
\begin{abstract}

Deep learning has made great strides for object detection in images, with popular models including Faster R-CNN, YOLO, and SSD. The detection accuracy and computational cost of object detection depend on the spatial resolution of an image, which may be constrained by both the camera and storage considerations.  Furthermore, original images are often compressed and uploaded to a remote server for object detection. Compression is often achieved by reducing either spatial or amplitude resolution or, at times, both, both of which have well-known effects on performance.  Detection accuracy also depends on the distance of the object of interest from the camera. Our work examines the impact of spatial and amplitude resolution, as well as object distance, on object detection accuracy and computational cost. As existing models are optimized for uncompressed  (or lightly compressed) images over a narrow range of spatial resolution, we develop a resolution-adaptive variant of  YOLOv5 (RA-YOLO), which varies the number of scales in the feature pyramid and detection head based on the spatial resolution of the input image. To train and evaluate this new method, we created a dataset of images with diverse spatial and amplitude resolutions by combining images from the TJU \cite{pang2020tju} and Eurocity \cite{braun2018eurocity} datasets and generating different resolutions by applying spatial resizing and compression. We first show that RA-YOLO achieves a good trade-off between detection accuracy and inference time over a large range of spatial resolutions. We then evaluate the impact of spatial and amplitude resolutions on object detection accuracy using the proposed RA-YOLO model. We demonstrate that the optimal spatial resolution that leads to the highest detection accuracy depends on the 'tolerated' image size (constrained by the available bandwidth or storage). We further assess the impact of the distance of an object to the camera on the detection accuracy and show that higher spatial resolution enables a greater detection range. These results provide important guidelines for choosing the image spatial resolution and compression settings predicated on available bandwidth, storage, desired inference time, and/or desired detection range, in practical applications.

%\keywords{Object Detection, Spatial Resolution, Image Compression, Accuracy/Speed Trade-offs, Detection Range}

\end{abstract}

%===========================================================
\section{Introduction}

  \begin{figure}[t]  \vspace{-.5em}
    \centering
    \includegraphics[width=9cm]{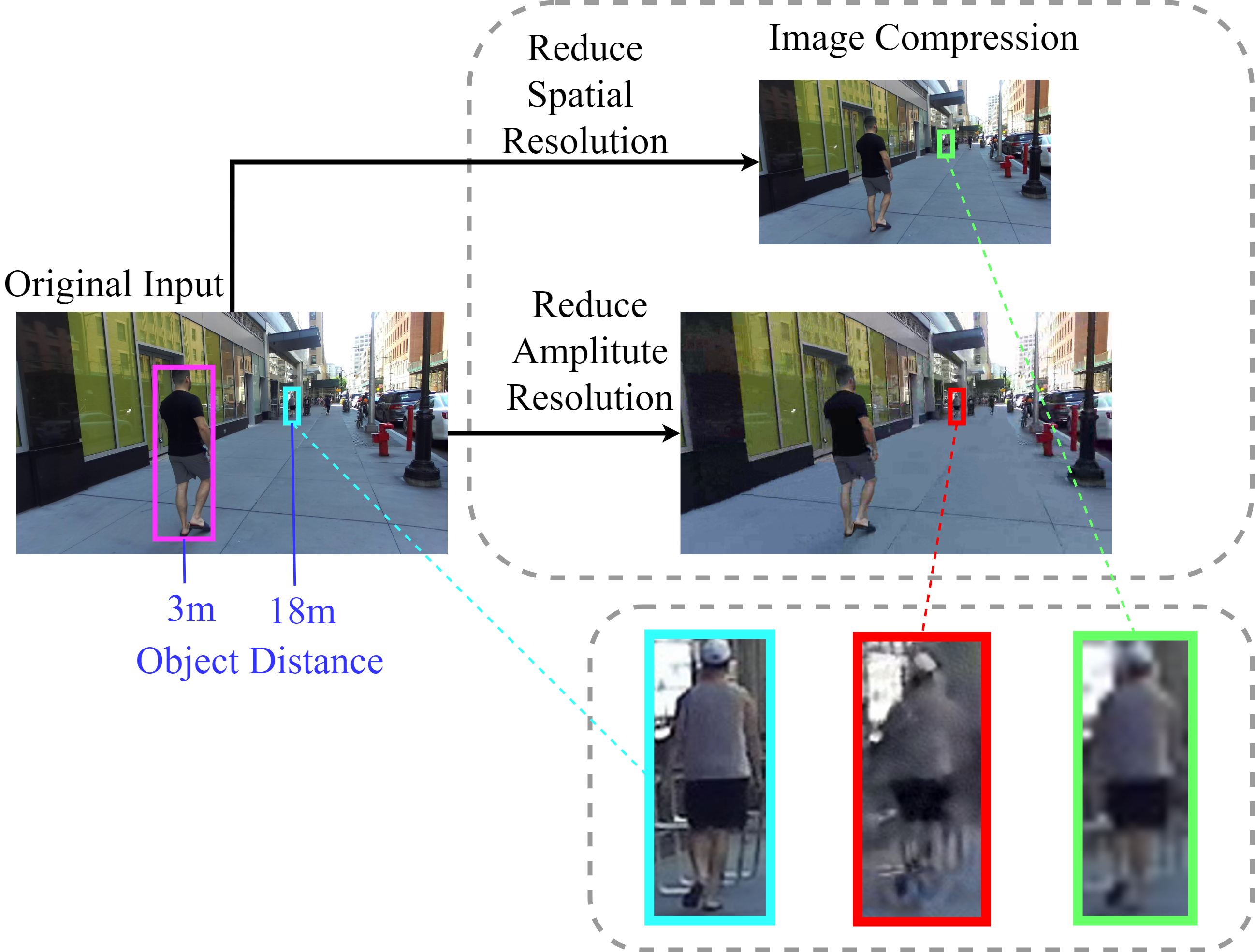}
    \vspace{-.5em}
    \caption{We show three main factors that influence the detection accuracy: spatial resolution, amplitude resolution and object distance to camera. The amplitude resolution is controlled by the quantization step size of the image coder. The original spacial resolution captured by the camera may be decreased to reduce the file size. }
    \label{fig3}
    \vspace{-1.5em}
\end{figure}

\vspace{-.5em}
Recently, learning-based data-driven methods have demonstrated great success in object detection, such as R-CNN \cite{girshick2014rich}, YOLO \cite{redmon2016you} and SSD \cite{liu2016ssd}. In real-world applications, images are inevitably captured with different spatial resolutions due to the differences in camera hardware and settings. Furthermore, the captured images are often compressed and uploaded to remote servers to speed up the computation.  The compression can be achieved by reducing spatial resolution and/or amplitude resolution (controlled by the quantization step size). The degree of compression depends on the available bandwidth.      
Inference time also plays an important role in real-time object detection. 
Although significant efforts have been devoted to improving the accuracy and speed of object detection, limited work, e.g. \cite{aqqa2019understanding}, explores how the image quality (controlled by the spatial and amplitude resolution) impacts object detection accuracy and inference time. Another important factor that affects the detection accuracy is the distance of the object to the camera. Close-by objects are easier to detect than far-away objects. In the literature, object detection accuracy is usually measured by mean accuracy for all objects regardless of their distances. In practical applications, such as autonomous driving \cite{zhu2019learning} or navigation for visually impaired persons \cite{li2019cross}, the detection range, defined by the maximum distance at 
which an object can be reliably detected, may be more relevant.
Based on the above observations, in this paper, as shown in Fig. \ref{fig3}, we aim to conduct studies to comprehensively evaluate how do image spatial resolution, amplitude resolution, and object distance impacts detection accuracy, and how these constraints affect inference time. 

One challenge in evaluating the impact of image resolution using existing object detection models is that they are not designed to operate on images over a large resolution range, rather separate models are optimized for different spatial resolutions. A common element among all object detectors is that it uses a multi-scale feature extractor (e.g. the feature pyramid in YOLO) to extract features at multiple scales, and detect objects at these scales using different detectors, and then combine all the detected bounding boxes using a non-maximum suppression step. This enables the detection of the same type of object appearing at different sizes, with fine-scale features facilitating the detection of small objects, and coarse-scale features enabling the recognition of large objects. Note that the size of an object (i.e. the bounding box size) depends on the distance of the object to the camera, as well as the image resolution. The necessary number of scales depends on the expected image resolution(s). For example, for low-resolution images (e.g. $640\times 640$), 3 scales may be sufficient (such as the widely used pre-trained YOLO model \cite{YOLO}; whereas for high-resolution images (e.g. $2560\times 2560$), 5 scales is necessary, in order to detect objects that appear from very small to very large in size (such as the pre-trained model in \cite{YOLO5scale}). Although the 5-scale model will also work well for low-resolution images, it forces unnecessary computation and extends inference time. Furthermore, in practice, it may not be efficient to store multiple models and switch them based on the input resolution. 

\begin{figure}[t] \vspace{-.5em}
    \centering
    \includegraphics[width=10cm]{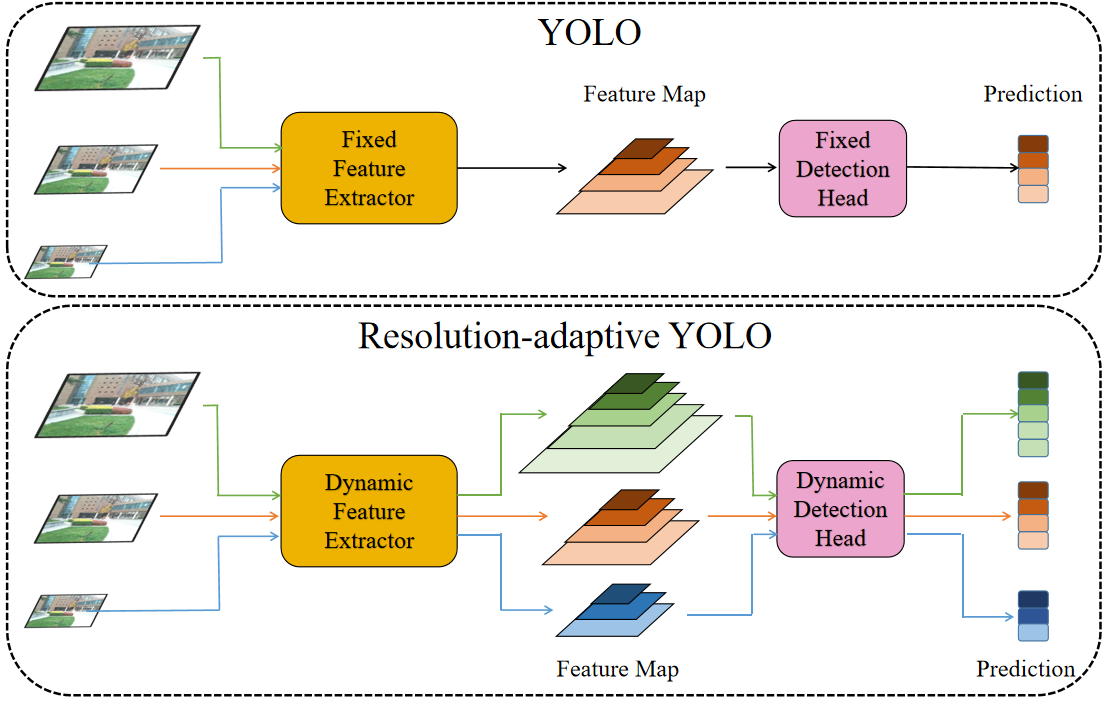}
    \vspace{-.5em}
    \caption{Comparison between current typical object detection methods and our proposed resolution adaptive model.}
    \label{fig1} 
    \vspace{-1.5em}
\end{figure}
%\vspace{-1em}

In this paper,  we first present a resolution-adaptive variant of the YOLOv5 architecture (called RA-YOLO), which varies the number of scales in the feature pyramid based on the spatial resolution of the input image, as shown in Fig. \ref{fig1}.  RA-YOLO contains five scales, however being mindful of inference time, some of the coarser scales can be dropped for lower-resolution images.  For images at  higher spatial resolutions, RA-YOLO utilizes all the scales.  For images at lower resolutions, RA-YOLO drops the last one or two scales to seek a balance between detection accuracy and computational complexity.
%Moreover, the model operates on the input image directly without resizing, which avoids information loss due to down-sampling. 
To improve the robustness of the model to the quantization artifact due to compression, we further refine the model by including training samples with varying amplitude resolutions. We then present evaluation results using RA-YOLO on images with different spatial and amplitude resolutions.  Our contribution is summarized as follows:

\begin{itemize}
\item We present a resolution-adaptive variant of the YOLOv5 model (RA-YOLO) that is designed to handle images of varying resolutions over a large range. It adapts the number of feature scales and detection heads based on the image resolution to achieve a good trade-off between detection accuracy and computational complexity. 

\item We conduct a comprehensive study to evaluate the impact of the image quality (spatial resolution and amplitude resolution) on detection accuracy.  We find that the detection accuracy decreases significantly when the spatial resolution reduces, as expected. Surprisingly, the detection accuracy is robust with respect to the amplitude resolution, until the amplitude resolution drops very low. We further show how to select the spatial resolution to maximize the detection accuracy for a given constraint on image file size.

\item We also evaluate the impact of the object distance on detection accuracy and show how that higher spatial resolution enables a longer detection range for pedestrians.

\item We investigate the trade-off between detection accuracy and inference speed under different image resolutions and the number of scales in the detector.

\item We contribute a new dataset that contains images compressed to different spatial and amplitude resolutions using the BPG codec. This dataset utilizes lightly compressed images from the TJU \cite{pang2020tju}, Eurocity \cite{braun2018eurocity}, and BDD \cite{yu2020bdd100k} datasets. The link to the dataset will be published alongside the paper. 
\end{itemize}

%Our prior work has studied the impact of spatial and amplitude resolution on the object detection accuracy and range, using a pre-trained YOLO model with 3 scales \cite{yuan2021network}. In that study, the highest spatial resolution considered is 2.2K (2208x1242), which had a very similar detection performance as for 1080P (1920x1080). The difference between 1080P and 720P (1280x720) is relatively small in part because we used a model with 3 scales only. By using RA-YOLO we can see more improvement as the spatial resolution increases.
%===========================================================

%===========================================================
\section{Resolution-Adaptive YOLO}

\subsection{Related Work in Deep Learning for Object Detection}

A great number of successful deep learning models such as YOLO \cite{redmon2016you}, SSD \cite{liu2016ssd} and Fast R-CNN \cite{girshick2015fast} have been developed to advance object detection. These pioneering works introduced a standard machine learning framework for object detection, which contains two basic modules: {\it Feature Extractor} and {\it Detection Head}, as shown in Fig. \ref{fig1}. The Feature Extractor module is often implemented with multiple groups of convolution layers, between which the feature maps are down-sampled, to extract feature representation at multiple scales. The Detection Head module is usually implemented with a few convolution layers or multiple layer perception (MLP) to predict the location (e.g. the object bounding box) of visual instances present in the image. Multiple detection heads are typically used, each acting on the feature representation at one scale. The bounding boxes detected at different scales are finally merged using a non-maximal suppression step.  A deep object detection model is often defined with two levels of parameters: the network hyper-parameter and network parameters. The network hyper-parameter, also referred as meta-architecture, defines the neural network architecture by specifying the number of network layers and the number of hidden neural nodes in each layer. The network parameters define the weights of neurons in the neural network. Many works (e.g., \cite{he2015spatial}\cite{girshick2015fast}\cite{ren2015faster}) have modified the standard pipeline with additional modules for better performance of object detection in different applications. Those works usually develop an advanced meta-architecture of deep neural networks whose neurons weights are often well-trained using large-scale datasets. After training, the detection model is deployed to test datasets without additional network hyper-parameter and parameters adjustments. However, existing detection models are not designed to operate on images over a large range of image resolutions, rather separate models are optimized for different spatial resolutions. The input images of varying resolutions are typically resized to the expected input size for the adopted model. In practice, it may not be efficient to store multiple models and switch them based on the input resolution. Further downsampling a high-resolution image to an expected low-resolution representation can remove the high-frequency details necessary to detect small objects and consequently reduce the detection accuracy.

\subsection{The Proposed RA-YOLO Model}
To overcome the above challenges, we have developed a resolution-adaptive variant of the YOLOv5 architecture (called RA-YOLO), which adaptively chooses the most effective meta-architecture of the detection network by varying the number of scales in the feature pyramid and consequently in the detection head based on the input image spatial resolution, as shown in Fig. \ref{fig1}. RA-YOLO is designed based on the observation that the apparent size (in terms of the size of the bounding box) of an object at a fixed distance to the camera is proportional to the image resolution.  For images at the higher spatial resolution, RA-YOLO utilizes more scales (corresponding to a deeper feature extractor and more detection heads) in order to detect objects that may appear from very small to very big.  For images at the lower resolution, RA-YOLO utilizes a shallow feature extractor and fewer detection heads to seek a balance between detection accuracy and computational complexity. Moreover, the model operates on the input image directly without resizing, which avoids information loss due to down-sampling. To improve the robustness of the model to the quantization artifact due to compression, we further refine the model by including training samples with varying amplitude resolutions.

\begin{figure}[t] \vspace{-.5em}
    \centering
    \includegraphics[width=12cm]{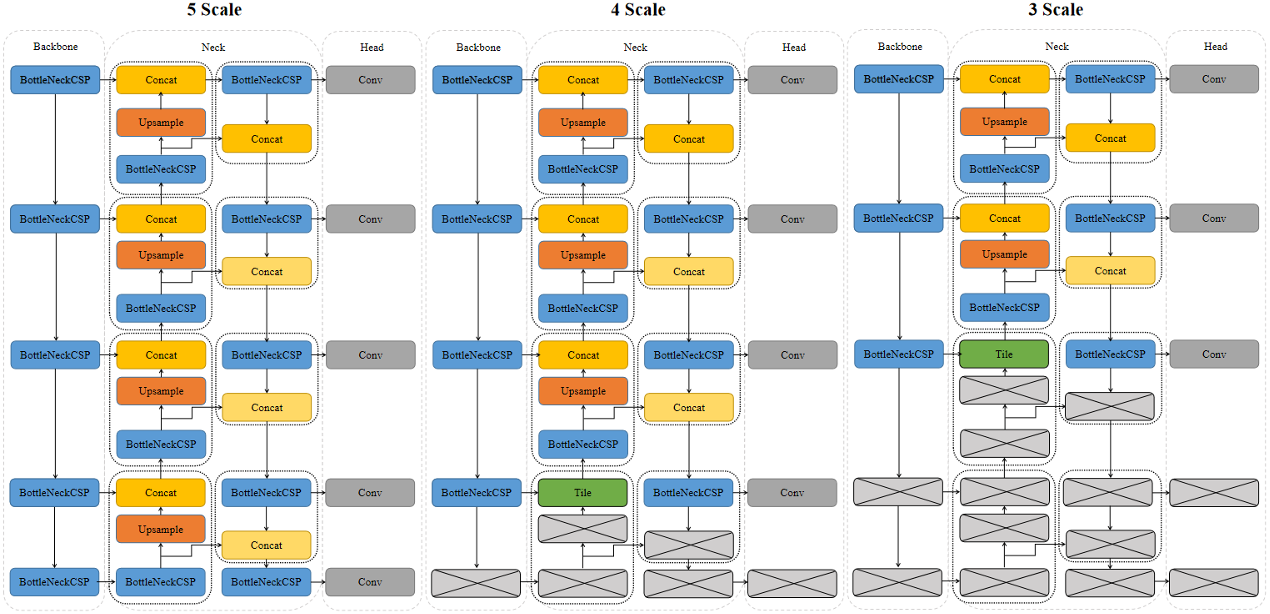}
    \vspace{-.5em}
    \caption{The RA-YOLO model architecture under different numbers of scales.}
    \label{fig:RA-YOLO} \vspace{-2em}
\end{figure}
%\vspace{-1em}

% \subsubsection{Implementation of RA-YOLO:}
\noindent{\textbf{Implementation of RA-YOLO:}}
We adopt the representative single-stage YOLO model \cite{redmon2018yolov3} as our meta-framework for object detection since it has the advantage of being faster than two-stage networks and still maintains promising accuracy. The proposed RA-YOLO is shown in Fig. \ref{fig:RA-YOLO}. The left-most column is the same as a YOLO model with 5 scales.
Given an input image, YOLO first generates multiple feature maps at different scales using a DarkNet backbone. The Feature Pyramid Network is utilized as the neck for feature aggregation at different scales. The aggregated features at each scale are then sent to a  detection head to extract candidate bounding boxes and furthermore  determine whether each candidate box contains an object and predicts the confidence score (or probability) for each possible object category. The non-maximum suppression method is finally applied to the output from all detection heads, to select the bounding box with the highest confidence score for an object category among overlapping candidates. The final output of the model includes detected bounding boxes, detected object category, and detection confidence. 
%We only select the prediction that exceeds a threshold on the intersection over union (IoU) between the detected box and the ground truth box and another threshold on the confidence score.  

With RA-YOLO, if the input image height $H$ is higher than a preset threshold (e.g. $H>H_5$ lines), the model will use all the 5 scales as illustrated in the left column of Fig.\ref{fig:RA-YOLO}. If the input resolution is intermediate (e.g.  $H_4 \leq H \leq H_5$), the model will skip the lowest scale (scale 5) as illustrated in the middle column. Here we replace the ``concatenate'' block in scale 4 with a ``tile' block, which duplicates the features from scale 4. Finally, if the input resolution is low (e.q. $H < H_4$), the model will further skip scale 4. We train the model using images with various resolutions and adapt the number of scales as indicated above for each training image. We use a batch size of 1 to accommodate varying image sizes per training sample. If a batch sample has low resolution, only the model weights for the top 3 scales are updated. On the other hand, if a batch sample has a high resolution, the model weights for all scales are updated. We use the same loss function as the original YOLO model, which includes the loss for bounding box coordinates and dimensions, the loss for the confidence of the object in the bounding box, and the classification loss for the detected objects \cite{YOLO}. 

%The entire loss is defined as follow:
%\begin{equation}
%\begin{split}
%L= 
%\mathcal{L}=\lambda_1 \mathcal{L}_{box} + \lambda_2 \mathcal{L}_{obj} + \lambda_3 %\mathcal{L}_{cls}
%\end{split}
%\end{equation}
%where $\lambda_1$, $\lambda_2$ and $\lambda_3$ are the pre-defined balance terms for bounding box loss, confidence loss, and classification loss.

%===========================================================
\section{Experiments}
\subsection{Model Setting and Running Environment} \label{sec:dataset}
For a fair comparison, we test our RA-YOLO model and all baseline models in a machine with an RTX 8000 GPU. For RA-YOLO, we set $H_4=810$ and $H_5=1620$.
%the base threshold $m$ as 810 and base scale $b$ as 3. For example, RA-YOLO will adaptively use 4 scales in the feature extractor and detection head for the images with resolutions ranging from 810 to 1620. 
That is, we consider three levels of resolution: low resolution (image height less than 810), middle resolution (image height between 810 and 1620), and high resolution (image height more than 1620). We use DarkNet \cite{redmon2018yolov3} as backbone and PANet \cite{liu2018path} as neck for the feature extractor. For determining the size of anchor boxes, we apply the autoanchor method to calculate the optimal anchor size for each scale based on the original image resolution.

 \subsection{Image Compression}
%Image quality is affected by spatial resolution, amplitude resolution. 
As shown in Fig. \ref{fig3},  images can be compressed by reducing the spatial resolution and/or amplitude resolution.  Given a particular spatial resolution, the amplitude resolution is controlled by the quantization step size of the image coder. We choose to use the BPG codec \cite{BPG}, which implements the HEVC intraframe coding method and has the leading coding efficiency among all standard codecs, significantly better than the more popular JPEG and JPEG2000 standards. It performs orthogonal transforms over variable-size blocks to exploit the  correlation among adjacent pixels and then applies quantization on resulting transform coefficients.   A larger quantization parameter  (QP) leads to smaller file sizes but more noticeable quantization artifacts.   
%For a given spatial resolution, the quality of the decompressed image is often quantified by the Peak Signal to Noise Ration (PSNR).

\begin{table}[h]
\centering \vspace{-3em}
\caption{Dataset Summary} \vspace{-1em}
\label{data_summary}
\begin{tabular}{llp{4.5cm}}\\ \hline
Dataset     & Spatial Resolution & Amplitude Resolution \\ \hline
\text{TJU Original}       & \text{2000P-4000P}         & QP=[0-51]          \\
\text{TJU Down2}   &  1000P - 2000P           & QP=[0-51]         \\
\text{TJU Down4}       & 500P - 1000P            & QP=[0-51]        \\ \hline
%\text{EuroCity Up2}        & 3840\times2048      & QP=[0-51]         \\ 
\text{EuroCity Original} & 1920$\times$1024      & QP=[0-51]         \\ 
\text{EuroCity Down1.42}        & 1350$\times$720      & QP=[0-51]         \\ \hline
BDD & 1280$\times$720      &  Original data are compressed at around 0.648 bits/pixel    \\ \hline    
\end{tabular} \vspace{-1.5em}
\end{table}     

\subsection{Dataset Preparation} \label{sec:dataset}
Standard benchmarks such as Imagenet \cite{deng2009imagenet} and COCO \cite{lin2014microsoft} have been well used in recent years for the object detection task. However, these mainstream benchmark datasets only consist of low-resolution images. Thus, we evaluate our proposed model on 3 object detection datasets with high, middle and low resolution: TJU \cite{pang2020tju}, EuroCity\cite{braun2018eurocity} and BDD\cite{yu2020bdd100k}. We choose two overlapping and representative categories (pedestrian, rider) in the three datasets as our object categories. A summary of all datasets is shown in  Table.\ref{data_summary}. Note that the rider category was annotated differently across the datasets. In TJU, the rider box includes both the person and the bicycle or motorcycle; whereas in EuroCity and BDD, the rider box only covers the person. Such inconsistency leads to lower detection performance for ``rider" than for ``pedestrian". The detection accuracy reported in this section is the mean average precision (mAP) over both categories. The AP for ``pedestrian'' is much higher.

% \vspace{-2mm}
% \subsubsection{TJU:} 
\noindent{\textbf{TJU:}}
The dataset contains 115,354 high-resolution images (90\% images have a resolution over 2000×2000 pixels) and 709,330 labeled objects in total with a large variance in size and appearance.

% \vspace{-2mm}
% \subsubsection{EuroCity:}
\noindent{\textbf{EuroCity:}}
The dataset contains a large number of middle-resolution images (1920×1024 pixels) in urban traffic scenes. The images for this dataset were collected on-board a moving vehicle in 31 cities of 12 European countries with over 238,200 person instances manually labeled in over 47,300 images. This dataset also has the object distance annotation.
 
%  \vspace{-2mm}
% \subsubsection{BDD:}
\noindent{\textbf{BDD:}}
BDD is a road scene benchmark consisting of 7,000 low-resolution images (1280x720) for training and 1,000 images for validation. The dataset possesses geographic, environmental, and weather diversity. For a fair comparison, we only use daylight images. The images in this dataset are already compressed with an average bit rate of 0.647 bits/pixel, with noticeable but not severe compression artifacts. %(Haoyang/Yixuan: correct) 
 
\begin{figure}[t]
    \vspace{-.5em}
    \centering
    \includegraphics[width=12cm]{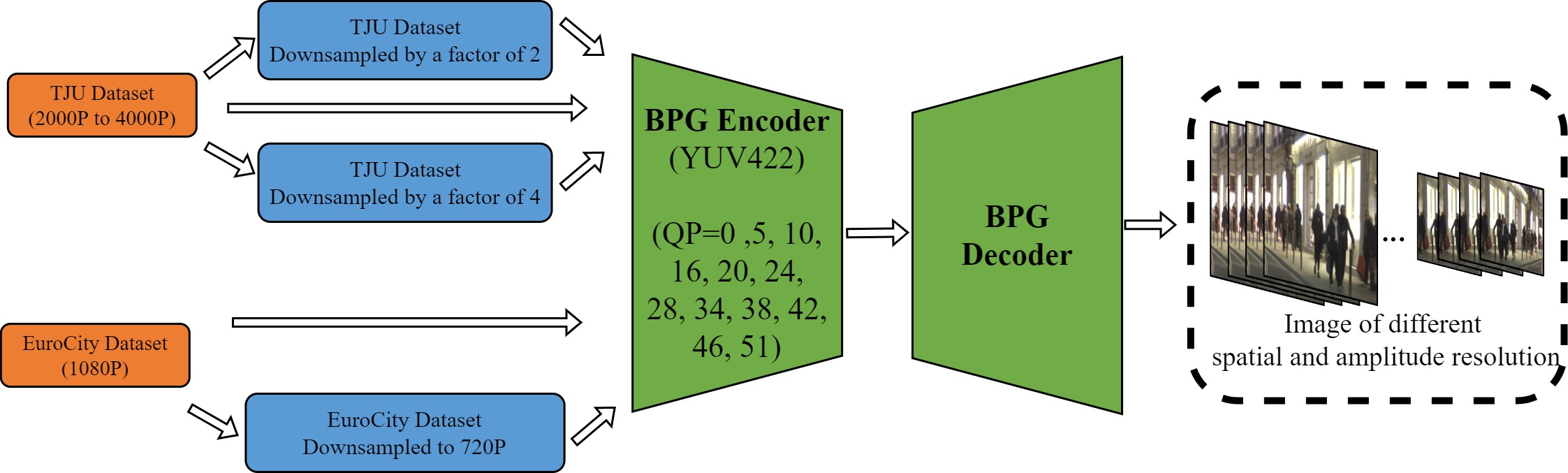}
    \vspace{-.5em}
    \caption{Data preparation for training and testing. First, all data from TJU and EuroCity datasets are converted to other spatial resolutions to have a wide range of spatial resolutions from 720P to 4000P. Data are then compressed by the BPG encoder to alter their amplitude resolution. 
    %This process simulates the compression step before data transfer from a camera sensor to a computation server where the chosen spatial and amplitude resolutions depend on available bandwidth.
    }
    \label{trans_compression}
    \vspace{-1.5em}
\end{figure}

% \subsubsection{Multi-Resolution Simulation for TJU and EuroCity Dataset:}
\noindent{\textbf{Generation of a Dataset with Mixed Spatial and Amplitude Resolutions:}}
In addition to the original images from the TJU and EuroCity datasets, we 
%upsampled the EuroCity images by a factor of 2 using Super-Resolution network Real-ESRGAN\cite{wang2021real} and 
downsampled the EuroCity and TJU images using FFmpeg\cite{FFmpeg} with the `bicubic' filter. We utilized the BPG (Better Portable Graphics) codec \cite{BPG} to compress and decompress the TJU and Eurocity datasets under different resolutions. The BDD data were compressed already at an average bpp of 0.647. This process is illustrated in Fig.\ref{trans_compression}.   Figure \ref{detec_result_dif_compression} provide sample images with different spatial and amplitude resolutions. We used subsets of this dataset for different studies reported in the remainder of this section.

\subsection{Evaluation Metric}
For evaluation of object detection performance, we use mean Average Precision (mAP) which is computed using the area under the precision-recall curve with a range of confidence thresholds, and a fixed threshold on the intersection of union (IOU) of 0.5 between the detected boxes and the ground truth boxes. 
%Given the thresholds on the IoU and confidence, all the detected boxes for a particular object category can be marked as true positive (TP), true negative (TN), false positive (FP), or false negative (FN), when compared with the ground truth box locations for this category. {\it Recall} and {\it Precision} are defined as:
%\begin{eqnarray*}
%\rm{Recall}=\frac{\rm \#TP}{\rm \#TP + \#FN}; \;\;\;
%\rm{Precision}=\frac{\rm \#TP}{\rm \#TP + \#FP}.
%\end{eqnarray*}

%YW: reviewers asked to remove definition of Recall and Precision

\begin{table}[t] \vspace{-.5em}
\small
\caption{ Speed/accuracy trade-offs under different spatial resolutions and number of scales. mAP in\%, inference time in ms, FLOP in Giga Flops. }

\scalebox{0.94}{
\begin{tabular}{c|ccc|ccc|ccc|ccc}
\hline
                 & \multicolumn{3}{c|}{YOLO(3)} & \multicolumn{3}{c|}{YOLO(4)} & \multicolumn{3}{c|}{YOLO(5)} & \multicolumn{3}{c}{RA-YOLO} \\ \hline
Resolution       & mAP      & Time    & FLOP     & mAP      & Time    & FLOP     & mAP      & Time    & FLOP     & mAP      & Time    & FLOP    \\ \hline
Low(BDD)         & 45.7    & 8.5     & 37.1     & 47.9    & 10.7    & 37.9     & 48.5    & 12.1    & 39.5     & 48.0    & 8.8     & 36.8    \\
Mid(EuroCity) & 72.7    & 17.8    & 75.0     & 74.4    & 19.2    & 76.7     & 75.2    & 22.5    & 84.2     & 74.9    & 19.1    & 76.4    \\
High(TJU)        & 74.9    & 71.3    & 423.1    & 76.7    & 75.8    & 432.5    & 77.5    & 81.0    & 450.3    & 77.7    & 80.6    & 450.3   \\
Average          & 64.4    & 32.5    & 178.4    & 66.3    & 35.5    & 182.4    & 67.0    & 38.4    & 191.3    & 66.9    & 36.1    & 187.9   \\ \hline
\end{tabular}\vspace{-1.5em}
}
\label{tradeoff}
\end{table}

\vspace{-1em} %adjust this along with \ref{table1 and detec_result_dif_compression}

\subsection{Speed vs. Accuracy Trade-Offs  Under Different Image Resolution and Network Scales}
\label{sec:trade-off}
% \subsubsection{Experimental Setting:} 
\noindent{\textbf{Experimental Setting:}}
In this section, we study the speed/accuracy trade-offs of our RA-YOLO model as well as baseline models using different numbers of network scales under different spatial resolution levels. We prepare training and testing samples  by sampling the images from the datasets of TJU Original, EuroCity Original, EuroCity Down1.42, and BDD as listed in Table \ref{data_summary}. For the high-resolution images, we collect images from the TJU  dataset with  resolution higher than 2000 lines. We randomly select 22K images, with 20k images used for model training and 2K used for testing. The medium-resolution images are chosen from the EuroCity dataset, with original resolution of  1920x1024.  We select 700 images from each city in the Eurocity training set as training data and 74 images from each city in the Eurocity validation set as  testing data. For the  low-resolution images, we sample 700 images from each city in the training set of EuroCity Down1.42 as  training data, and we  randomly select  2k images from the BDD  dataset as testing data. All training data collected at three different resolutions above are then combined to form the final mixed-resolution data for model training.  For testing, we evaluate the model performance on testing images with different resolutions separately.  Note that for this study, the training and testing data are not compressed (except the BDD testing data, which are compressed).

Recall that the RA-YOLO model takes  an image in its original resolution and adaptively determines the number of scales in the feature extractor and detection heads according to the image resolution.
We compare the RA-YOLO model with three baseline models: YOLO(3), YOLO(4), YOLO(5), which are standard YOLO models with 3, 4, and 5 scales, respectively. 
For example, YOLO(5) denotes that the YOLOv5 model that uses 5 scales for the feature extractor and detection head. Each baseline model is trained using the same mixed-resolution training set, and uses the original resolution of each image directly (i.e. without resizing) both during model training and testing.

\noindent{\textbf{Results:}}
In Table \ref{tradeoff}, we compare RA-YOLO with the three baseline models in terms of detection accuracy (in terms of mAP) and complexity (in terms of inference time and FLOP count per image) for testing images at different spatial resolutions.
%To measure the model complexity, we also report the FLOPs as the total number of computations of a model per image . 
%( I think here we should emphasize that for a input resolution lower than the highest, RA-YOLO has a complexity lower than YOLO(5), but achieves similar detection accuracy. Furthermore, RA-YOLO has better detection accuracy than YOLO(3) for low res and than YOLO(4) for mid-res, even though the same number of scales were used. Should explain why. Perhaps it is because RA-YOLO training strategy. Ex: YOLO(3) trained using all images, and anchors for all scales were set based on all these images. Whereas RA-YOLO's first 3 scales anchor were chosen primarily for low-resolution images?)
The results in Table \ref{tradeoff} show that increasing spatial resolution can greatly improve the detection accuracy  either with a fixed network scale or adaptive scale. There is a substantial increase in mAP from the low resolution to the medium resolution.  When the resolution changed from medium to high, there is still  significant accuracy gain. We should note that the low-resolution BDD data was not included in the training set and was compressed, and hence had far worse accuracy than the other  low-resolution EuroCity Down 1.42 images. However, Fig. \ref{tjumap50vsimgs} and Fig. \ref{euromap50vsimgs} show that even between Eurocity Original and Eurocity Down1.42, or between TJU Down2 and TJU Down4, there is a large gap in the detection detection accuracy between these two resolutions.

In addition, the experimental results also show that scale selection is critical for object detection. Generally, more scales will give more accurate detection. However, using RA-YOLO, which uses fewer scales for lower resolution images, we can achieve very similar mAP as always using 5 scales, but with faster computation speed. As shown in Table \ref{tradeoff}, for low-resolution images, our RA-YOLO (which uses 3 scales for this resolution) achieves better detection accuracy and similar inference time and FLOPs in comparison with YOLO(3). For the medium-resolution image,  RA-YOLO (using 4 scales), achieves a slightly better accuracy and similar speed and FLOPs  compared to YOLO(4). Finally, for the high-resolution images,  the accuracy of RA-YOLO (using 5 scales) is on par with YOLO(5) and has similar inference time and FLOPs. On average over all three resolutions, using more scales always provides more accurate detection, but at the expense of computation speed and complexity.  RA-YOLO achieves similar average detection accuracy as YOLO(5) with faster speed and lower complexity, albeit the reduction in complexity is limited. Overall, we  conclude that the RA-YOLO has a great speed and accuracy trade-off for images at various spatial resolutions, compared to using the standard YOLO model with a fixed number of scales.

It is expected that for the same image resolution and using the same number of scales, the YOLO model and the RA-YOLO should have similar running time and complexity. It was a bit surprising that RA-YOLO achieves slightly higher detection accuracy.  
We suspect this is because the training strategy for RA-YOLO optimizes the performance for each resolution by using the corresponding number of scales. For example, YOLO(3) is trained to optimize the average performance over training images with both low and high resolutions  using 3 scales always, so the performance for low-resolution images may be a bit compromised, so as not to degrade the performance for high-resolution images. On the other hand,  RA-YOLO is trained to optimize the performance over training images with the number of scales appropriate for their resolutions. The anchor sizes chosen at each scale with RA-YOLO may also be appropriate for that scale.

%The superior performance of Model C and Model F suggests that for images at the higher spatial resolution, the detection model generally requires more scales, which corresponds to a deeper feature extractor and more detection heads, in order to detect objects that may appear from small to large in size. Different from Model D, Model E, and Model F,  RA-YOLO adaptively chooses the most effective meta-architecture of the detection network of appropriate scales in the feature pyramid based on the input image spatial resolution. RA-YOLO dynamically utilizes more scales in order to detect objects that may appear from very small to very big. For images at the lower resolution, RA-YOLO utilizes a shallow feature extractor and fewer detection heads to seek a balance between detection accuracy and computational complexity. In the experiment, RA-YOLO achieves a comparative performance with Model F which has a fixed scale of 5 in its network. 

%Haoyang/Yixuan   
\vspace{-1em} %adjust this along with \ref{table1 and detec_result_dif_compression}
\subsection{Impact of Spatial and Amplitude Resolution }
\label{sec:refine}
% \subsubsection{Experimental Setting:}
\noindent{\textbf{Experimental Setting:}}
%In this section, we conduct experiments on the combined dataset containing both TJU and EuroCity images with different spatial and amplitude resolutions to examine the impact of amplitude resolution on the object detection accuracy under different spatial resolutions. 
%We applied our Multi-resolution Object Detection Model (RA-YOLO) on decompressed TJU(original, down by 2 and 4) and EuroCity(up, original and down-sampled) to detect pedestrians and riders. 
In this section, we first evaluate the detection accuracy of the RA-YOLO model trained using the uncompressed mixed-resolution data, described in Sec. \ref{sec:trade-off} on compressed testing images in TJU and EuroCity with different spatial and amplitude  resolutions. 
Because Eurocity images typically have fewer instances of objects of interest, we randomly choose 1000 images from TJU and 2163 images from EuroCity, to maintain similar  number of object instances from these two datasets in the testing set. In order to improve the model performance on the low amplitude resolution images, we further finetuned the RA-YOLO model with 9000 decompressed images evenly chosen from TJU(Original), EuroCity(Original), Eurocity(Down) training set with QP from 34 to 51.

\begin{figure}[t]
\centering
\begin{subfigure}{0.495\textwidth}
    \includegraphics[width=\textwidth]{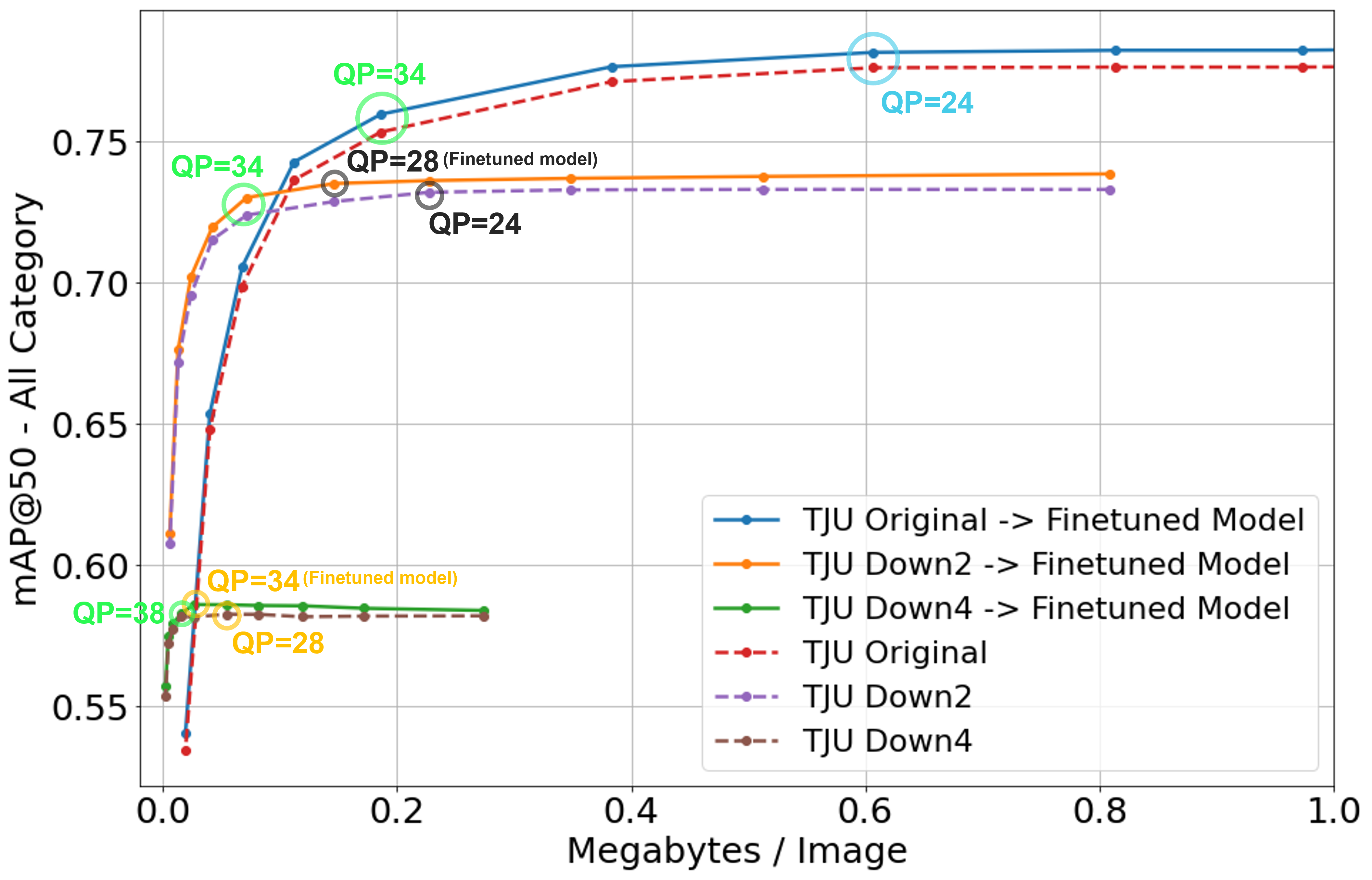}
    \caption{mAP@50 vs Image Size for TJU}
    \label{tjumap50vsimgs}
\end{subfigure}
\hfill
\begin{subfigure}{0.49\textwidth}
    \includegraphics[width=\textwidth]{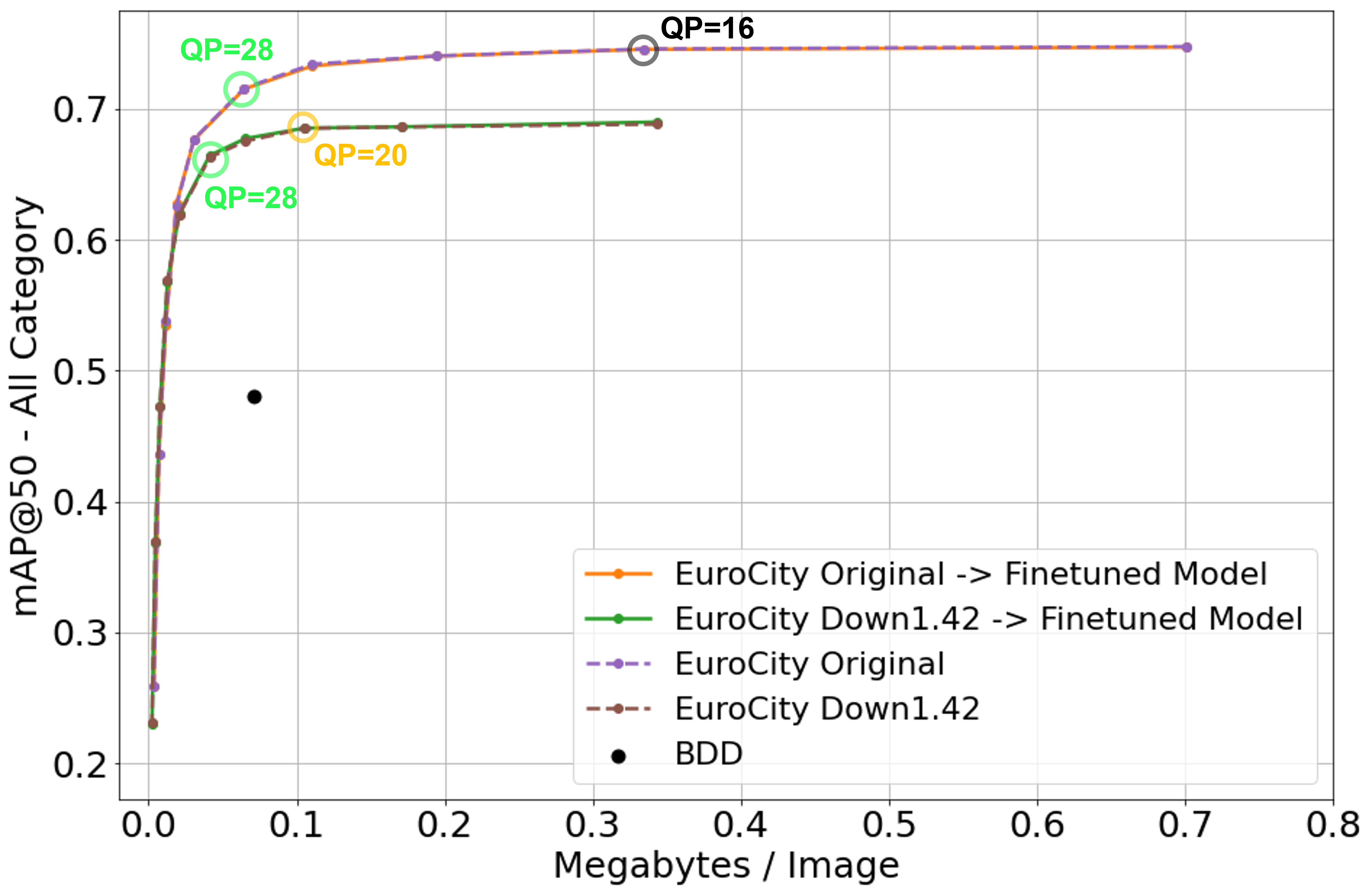}
    \caption{mAP@50 vs Image Size for EuroCity}
    \label{euromap50vsimgs}
\end{subfigure}
\hfill
\begin{subfigure}{0.49\textwidth}
    \includegraphics[width=\textwidth]{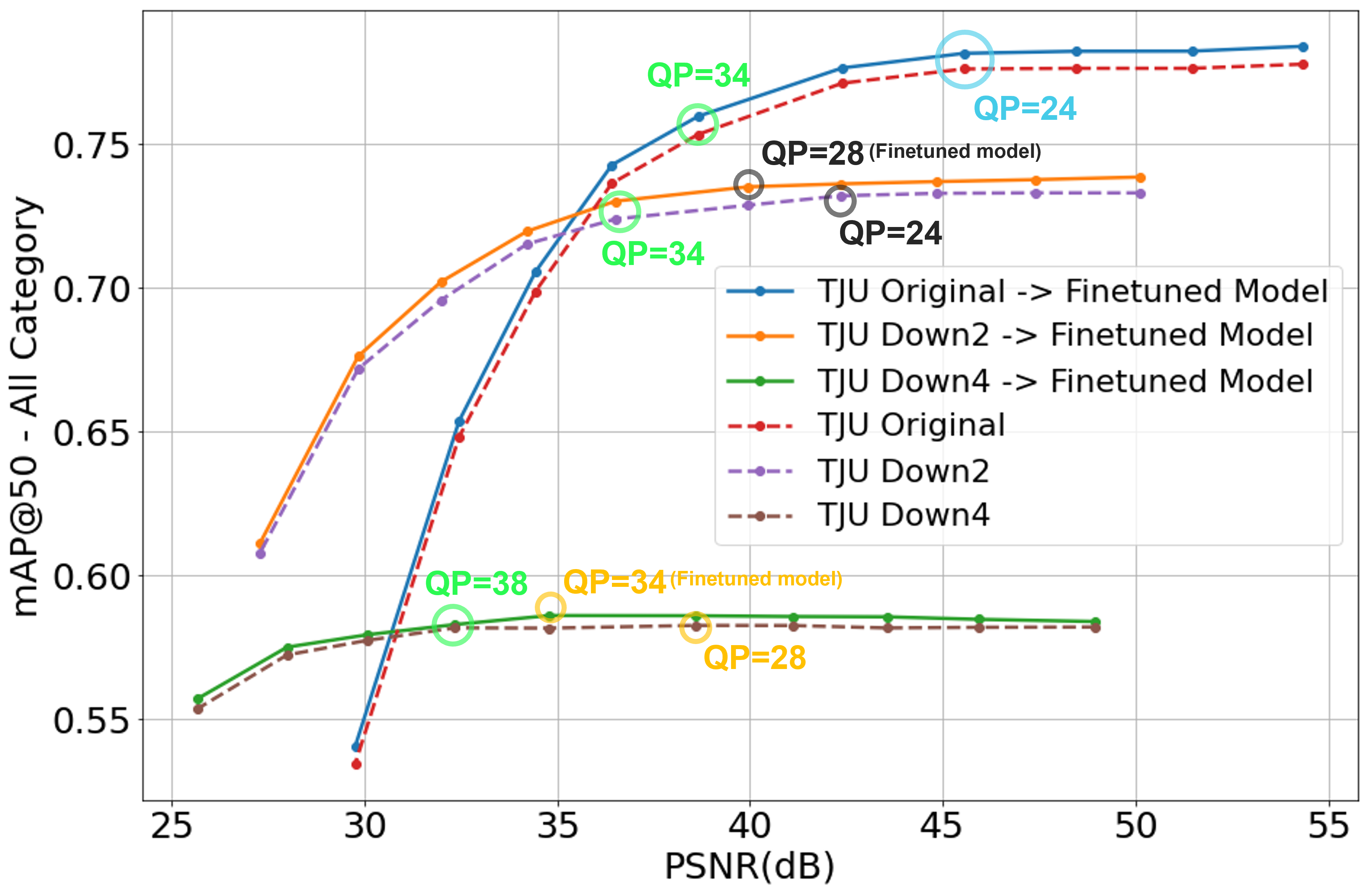}
    \caption{mAP@50 vs PSNR for TJU}
    \label{tjumap50vspsnr}
\end{subfigure}
\hfill
\begin{subfigure}{0.49\textwidth}
    \includegraphics[width=\textwidth]{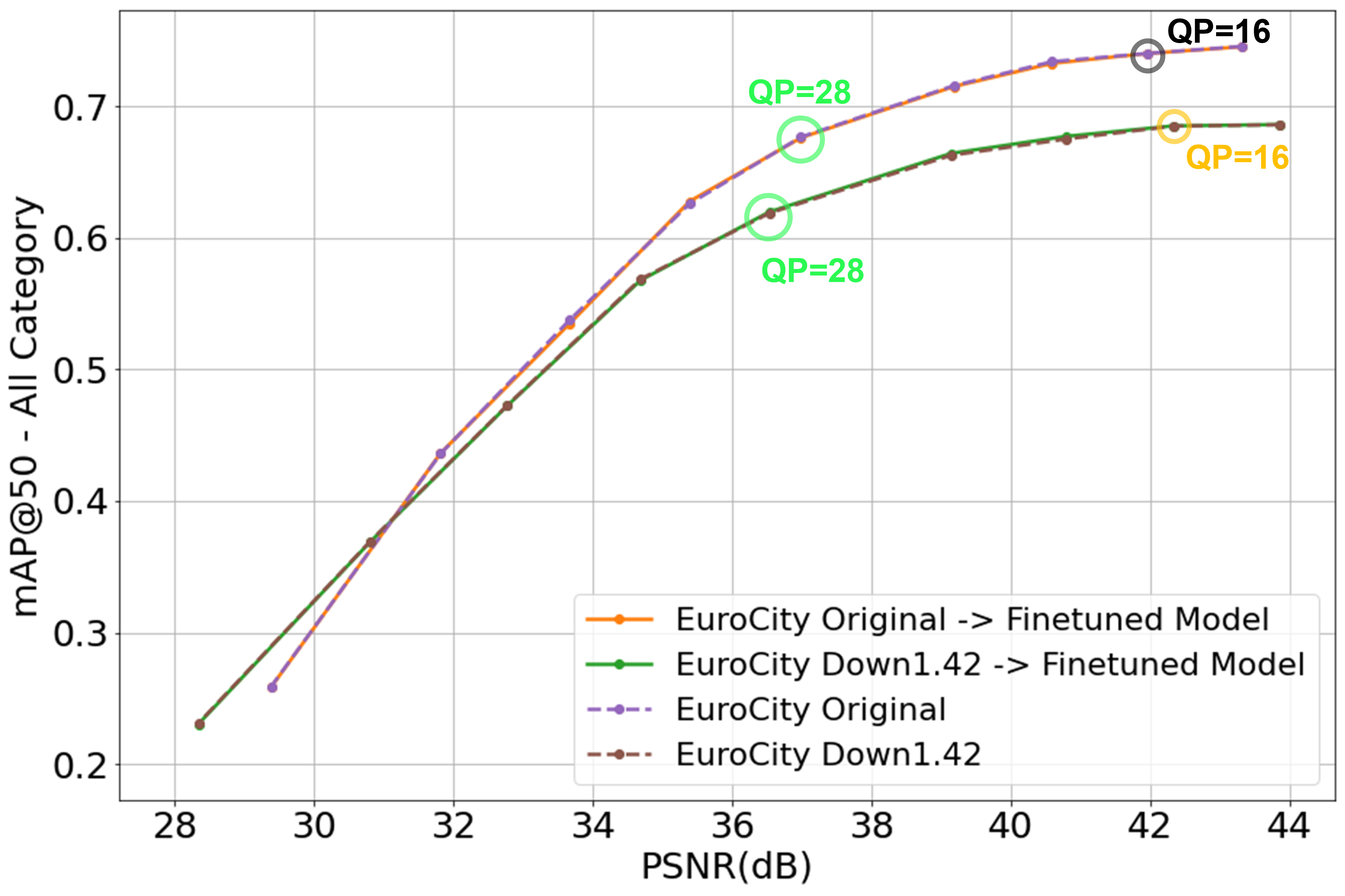}
    \caption{mAP@50 vs PSNR for EuroCity}
    \label{euromap50vspsnr}
\end{subfigure}

\caption{Detection accuracy vs (a,b) Image Size and (c,d) PSNR under different spatial resolutions.}
\label{map_psnr_imgs_curve}\vspace{-2em}
\end{figure}

% \begin{figure}[t]
%   % \centering \vspace{-2em}
%     \begin{minipage}[t]{0.48\textwidth}
%     \includegraphics[scale=0.185]{tju_map_mb_all.png}
%     \end{minipage}
%     \begin{minipage}[t]{0.48\textwidth}
%     \includegraphics[scale=0.185]{euro_map_mb_all.png}
    
%     \end{minipage} \vspace{-1.em}
%     \caption{mean Average Precision at IOU=0.5 vs Image Size (Megabytes)}
%     \label{mAP@50_all} \vspace{-1.5em}
% \end{figure}
% \begin{figure}[h]
%   % \centering \vspace{-2em}
%     \begin{minipage}[t]{0.48\textwidth}
%     \includegraphics[scale=0.2]{tju_psnr_map.png}
%     \end{minipage}
%     \begin{minipage}[t]{0.48\textwidth}
%     \includegraphics[scale=0.2]{euro_psnr_map.png}
%     \end{minipage} \vspace{-1.em}
%     \caption{mean Average Precision at IOU=0.5 vs PSNR}
%     \label{mAP_psnr} \vspace{-1.5em}
% \end{figure}
%Haoyang and Yixuan: please filling in necessary information below
% \subsubsection{Results:}

\noindent{\textbf{Results:}}
The dash curves in Fig. \ref{tjumap50vsimgs} and Fig. \ref{euromap50vsimgs} are results of the model trained using uncompressed images  (original images and downsampled ones). As expected, the spatial resolution significantly affects the detection accuracy, with higher resolution typically giving much higher mAP.  Somewhat surprisingly, compression does not reduce the detection accuracy until the  images are reduced by half (corresponding to QP of 24-28 for TJU and 16 for Eurocity). When QP is further increased, the detection accuracy starts to gradually drop. This means that our object detector is robust to  modest levels of quantization.
%It also suggests that for a given spatial resolution, we can compress the image using a relatively high QP up to 28-34 (consequently reducing the image size) without significantly affecting the detection accuracy. 
When QP exceeds a certain threshold (34 for TJU, 28 for Eurocity), the quantization artifact makes it harder to recognize the object, leading to a rapid decrease in the detection performance.  

\begin{figure}[t]
    \centering
    \includegraphics[width=12cm]{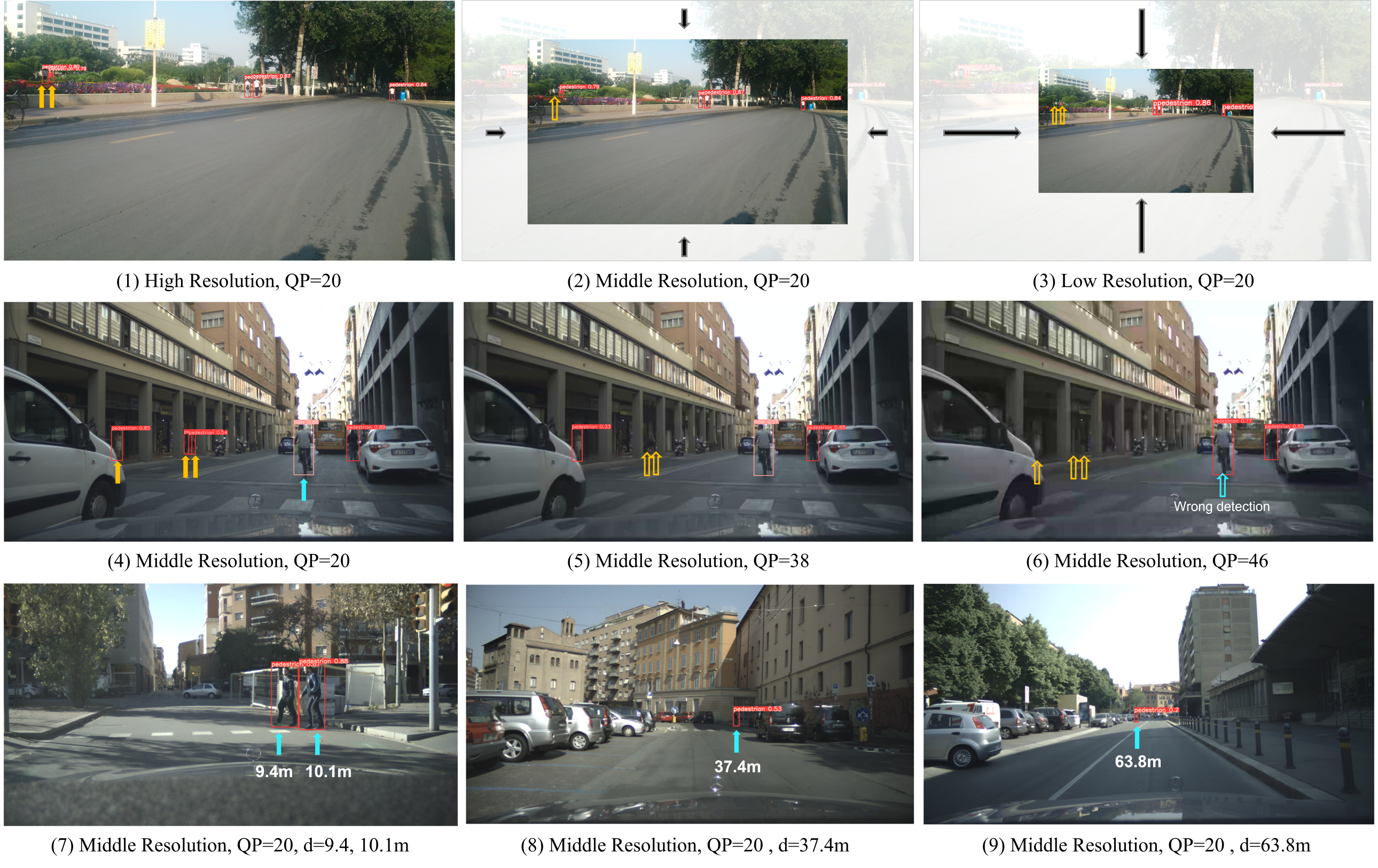}
    \vspace{-.5em}
    \caption{Sample detection results for images under different  spatial resolution, amplitude resolution, and object distance. %This figure visually shows how spatial resolution, amplitude resolution, and distance affect the number of the object detected, hence the detection accuracy. 
    The first row shows the effect of reducing the spatial resolution. 
    The second row shows the effect to reducing the amplitude resolution (by increasing QP).
    The third row shows the difficulty when the object is  farther away. 
    Dashed yellow arrows show the objects that were miss-detected, compared to the left image. Dashed blue arrows show the objects that were misclassified.
    %From (1) to (3) and (2) to (4), the spatial resolution decreases by the factor of 1.42. From (1) to (2) and (3) to (4), the QP increases from 20 to 38. Then the number of the object detected also decrease as decreasing spatial resolution or increasing QP. Note that even at a higher QP=38, increasing the spatial resolution can still help increasing the number of the objects detected. Compared with (5) and (6), we can observe that the object at relatively far away distance is hard to detect, thus the detection accuracy will drop as increasing object distance.
    }
    \label{detec_result_dif_compression} \vspace{-1em}
\end{figure}

The solid curves show the results using the fine-tuned model with additional examples with heavy quantization.  Overall, such fine-tuning achieves only slight improvement, roughly 0.01 in mAP. Surprisingly, fine-tuning using heavily quantized samples (QP from 34 to 51)  not only improves detection accuracy for images at those quantization levels but also lifted the performance of images at lower quantization levels. We suspect that such additional training examples helped the trained model to be more robust to not only quantization noise but also other artifacts such as those due to low lighting and motion blur.  Top two rows in Fig. \ref{detec_result_dif_compression} show sample images and their detection results   under different spatial and amplitude resolutions.

Because PSNR is commonly used to assess the image quality under the same spatial resolution and is a good quantitative metric to evaluate the quantization-induced artifacts, Fig. \ref{tjumap50vspsnr} and Fig. \ref{euromap50vspsnr} show how does the detection accuracy correlates with PSNR (or image quality). Here we clearly see that the mAP saturates after PSNR reaches a relatively high level, but drops sharply when PSNR decreases below a relatively low level. The particular thresholds depend on the spatial resolution and scene content. This is an interesting result as it tells us that one can compress images in amplitude  quite substantially without affecting the detection accuracy, even when the images themselves  have  degraded quality. 

%(YW: perhaps it is more interesting to include figures of PSNR vs image size, to show that even when PSRN drops as image size reduces, the AP keeps similar until the image size (or PSNR or QP) reduces significantly.)

Importantly, Fig. \ref{tjumap50vsimgs} and Fig. \ref{euromap50vsimgs} provide a guideline on how to choose the spatial resolution for a given constraint on the image file size. It shows that, above a certain image size constraint, it is better to compress a high-resolution image, than to use an uncompressed or lightly compressed low-resolution image. %Although the original high-resolution image leads to higher detection accuracy for relatively large file sizes,  
Only when the average file size needs to be below a certain threshold, we should switch to a lower resolution. For example, for TJU images, we should downsample the images by a factor of 2 when the network bandwidth limits the image size to be below 0.1 Megabytes. For EuroCity, even though the down2 resolution gives similar mAP at a very small image size, using the down2 resolution can reduce the computation cost without sacrificing the detection accuracy.

\subsection{Impact of the Object Distance}
% \subsubsection{Experimental Setting:}
\noindent{\textbf{Experimental Setting:}}
In this section, we evaluate the impact of the object distance on the detection accuracy under different spatial resolutions. Because only the EuroCity dataset provides distance information, we perform this evaluation using only the original images in the EuroCity dataset and their down-sampled versions. We use the RA-YOLO model refined using compressed data as discussed in Sec.\ref{sec:refine}. Although we do not include compressed images for the testing, we expect the results to be similar for QP up to 20. The object distance in EuroCity data covers a wide range from 0m to 80m. We quantize the distance to 8 bins of 10 m intervals and measure the detection accuracy. For this evaluation, we associate the predicted box with the ground truth box that has the highest IOU with the detected box and evaluates the recall of all ground truth boxes within each distance bin under a chosen confidence threshold.
%Since the predicted bounding box does not have the distance information, we set the distance  the real distance of a groundtruth box with the highest IOU with the detected box.  to each detected bounding box: for each detected box in one image sample, we calculated their IOU with each ground truth box and matched each detected box with a distance corresponding to the ground truth box with a highest IOU. In this rule, some false positive detected boxes with IOU=0 removed before computing metrics, which will lead to a higher mAP score. Therefore, we only evaluate recall(true positive rate) at each distance bins. 
For a fair comparison between each spatial resolution, the confidence score is set to 0.36 and 0.23 respectively for the original and down2 resolution to maintain the same precision of 0.82. Because the ``rider'' category does not have sufficient instances in certain distance bins, we only perform this evaluation on the ``pedestrian'' category.

%Yixuan/Haoyang: please change the figure so (a) is recall vs. distance, (b) is the detection range
\begin{figure}[h]
\centering 
\includegraphics[width=12cm]{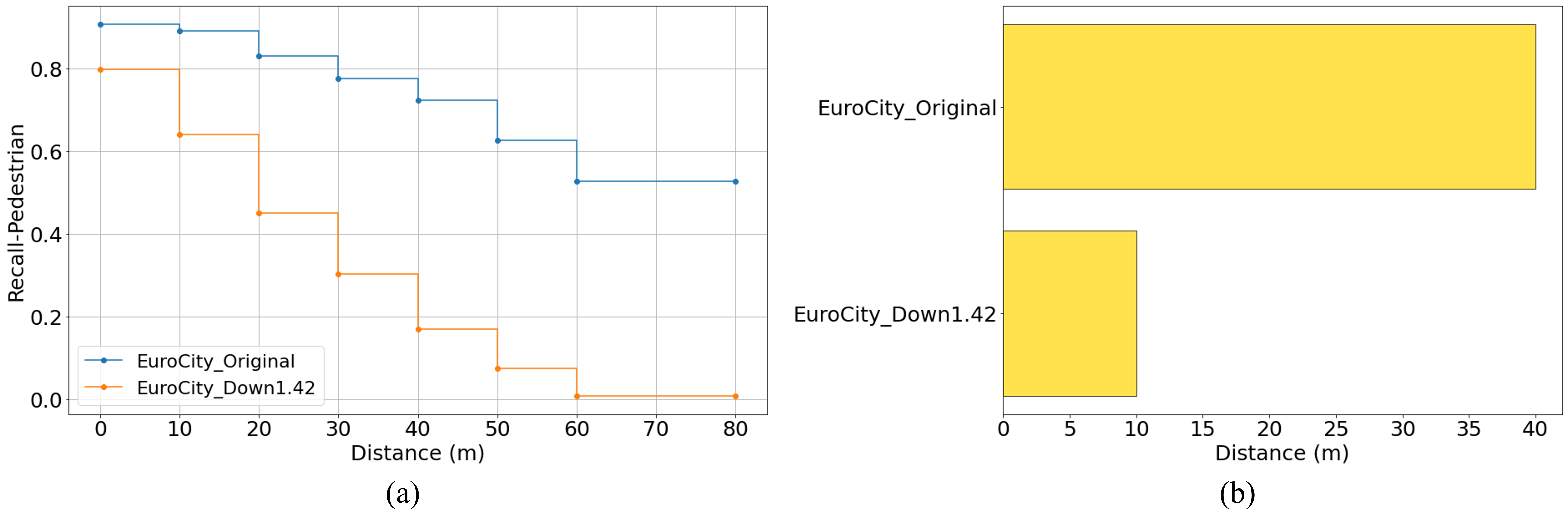} \vspace{-1em}
\caption{(a) Recall vs distance under the  precision of 0.82  for Pedestrian; (b) The detection range to achieve recall of 0.77 and precision of 0.8. }
\label{recall_distance} \vspace{-2em}
\end{figure}
% \subsubsection{Results:} 

\noindent{\textbf{Results:}}
 Fig. \ref{recall_distance}(a) reveals how the recall decreases as the object distance increase under different spatial resolution using RA-YOLO. 
%When the distance is very close (0-10m), our RA-YOLO performs very well under the original spatial resolution. 
Notice that at the short distance, even the down1.42 images enjoy a descent detection accuracy for the pedestrian category, much higher than the average detection accuracy shown in Fig. \ref{tjumap50vsimgs} for both pedestrian and rider.
As the distance increases, the recall decreases and the drop is slower for the higher resolution. This is as expected since the same object in the higher resolution image is rendered with more pixels or correspondingly higher clarity than that in the lower resolution.
Fig. \ref{detec_result_dif_compression} presents example object detection results under different object distances in the third row. In this example, a person that is 63.8m away was detected successfully, even though it was hard to discern with our naked eye. 

For practical applications, we want to know what is the farthest distance at which an object of interest can be detected reliably. 
Fig. \ref{recall_distance}(b) shows the detection range defined as the farthest distance when the precision is 0.8 and recall is 0.77.
We can see that the original spatial resolution with 1024 lines enables reliable detection up to 40 m while reducing the resolution to 720 lines  decreases the range to 10 m only. Therefore high resolution is critical for applications such as autonomous driving.

\vspace{-0.5em}
\section{Conclusion}
\vspace{-0.5em}
In this paper, we provide a comprehensive study to evaluate how image quality (in terms of spatial resolution and amplitude resolution) and the distance of objects in the image affect detection accuracy. Given that existing object detection models are generally not optimized for images over a large range of spatial resolutions, we propose a new variant of the YOLOv5 model, which adaptively selects the most effective meta-architecture (including the number of feature scales and correspondingly the number of detection heads)  according to the spatial resolution of the input image. We show that this resolution-adaptive model (RA-YOLO) provides a good trade-off between the detection accuracy and computational complexity, reaching a similar average AP over a large range of spatial resolutions as compared to the generic YOLOv5, while enjoying a faster inference speed. 
Using RA-YOLO on a mixed resolution dataset that we created from 3 open datasets of different spatial resolutions, we evaluated the impact of spatial resolution and compression (affecting the amplitude resolution) on the detection accuracy. We show that high spatial resolution is critical for achieving high average detection accuracy over a large range of object sizes (which is  affected by both the physical size of the object and the distance of the object from the camera). For example, for the TJU data, the mAP decreased from 77.7\% to 58.2\%, when we reduced the original high-resolution images (2000P-4000P) by a factor of 4 (500P-1000P). However, when the original high-resolution images need to be delivered to a remote edge server for computing, the images need to be compressed to fit bandwidth constraints.  We demonstrate that the RA-YOLO model is fairly robust to compression artifacts until the amplitude resolution is significantly reduced. Further refinement of the model using compressed images will help boost the detection performance.   The optimal spatial resolution that leads to the highest detection accuracy depends on the desired image size (constrained by the available bandwidth). 
We further show that using  high-resolution images can substantially extend the reliable detection range for pedestrians: from 10m using 720P resolution to 40m using 1024P resolution. We expect that similar results hold for riders, and that using higher resolution images (e.g. greater than 2000P) can further extend the range.  This suggests that using high-resolution images is critically important for applications that require reliable detection of objects that are far away, such as autonomous driving or smart health applications.

%===========================================================
\bibliographystyle{splncs}
\bibliography{egbib}

%this would normally be the end of your paper, but you may also have an appendix
%within the given limit of number of pages
\end{document}